\documentclass[sigconf]{acmart}

\usepackage{bbm}
\usepackage{multirow}
\usepackage{hyperref}
\usepackage{color}
\usepackage{amsmath}
\usepackage{xcolor}
\usepackage{listings, xcolor}
\usepackage{colortbl}
\usepackage[normalem]{ulem}
\usepackage{xspace}
\usepackage{enumitem}

\definecolor{lightblue}{rgb}{0.85, 0.95, 1.0}

\usepackage{pgfplots}
\usepackage{pifont}

\usepackage{tcolorbox}
\usepackage{adjustbox}

\usepackage{bbm}
\usepackage{multirow}
\usepackage{hyperref}
\usepackage{color}
\usepackage{amsmath}
\usepackage{xcolor}
\usepackage{colortbl}
\usepackage[normalem]{ulem}
\usepackage{enumitem}
\usepackage{tcolorbox}
\usepackage{adjustbox}
\usepackage{subfigure}
\usepackage{xspace}
\usepackage{balance}
\usepackage{hyperref}

\usepackage{amssymb}

\makeatletter
\def\@ACM@checkaffil{
    \if@ACM@instpresent\else
    \ClassWarningNoLine{\@classname}{No institution present for an affiliation}%
    \fi
    \if@ACM@citypresent\else
    \ClassWarningNoLine{\@classname}{No city present for an affiliation}%
    \fi
    \if@ACM@countrypresent\else
        \ClassWarningNoLine{\@classname}{No country present for an affiliation}%
    \fi
}
\makeatother

\AtBeginDocument{%
  }

\copyrightyear{2024}
\acmYear{2024}
\setcopyright{acmlicensed}\acmConference[CIKM '24]{Proceedings of the 33rd ACM International Conference on Information and Knowledge Management}{October 21--25, 2024}{Boise, ID, USA}
\acmBooktitle{Proceedings of the 33rd ACM International Conference on Information and Knowledge Management (CIKM '24), October 21--25, 2024, Boise, ID, USA}
\acmDOI{10.1145/3627673.3679673}
\acmISBN{979-8-4007-0436-9/24/10}

\begin{document}

\title{Editing Factual Knowledge and Explanatory Ability of Medical Large Language Models}


\author{Derong Xu}
\affiliation{%
  \institution{University of Science and Technology of China, \\\& City University of Hong Kong}
  \city{Hefei}
  \country{China}
}
\email{derongxu@mail.ustc.edu.cn}

\author{Ziheng Zhang }
\affiliation{%
  \institution{Jarvis Research Center, \\ Tencent YouTu Lab}
  \city{Shenzhen}
  \country{China}
}
\email{zihengzhang@tencent.com}

\author{Zhihong Zhu }
\affiliation{%
  \institution{Peking University}
  \city{Beijing}
  \country{China}
}
\email{zhihongzhu@stu.pku.edu.cn}

\author{Zhenxi Lin}
\affiliation{%
  \institution{Jarvis Research Center, \\ Tencent YouTu Lab}
  \city{Shenzhen}
  \country{China}
}
\email{chalerislin@tencent.com}

\author{Qidong Liu }
\affiliation{%
  \institution{Xi'an Jiaotong University, \\City University of Hong Kong}
  \city{Xi'an}
  \country{China}
}
\email{liuqidong@stu.xjtu.edu.cn}

\author{Xian Wu \dag}
\thanks{\dag \ \text{Corresponding authors}}
\affiliation{%
  \institution{Jarvis Research Center, \\ Tencent YouTu Lab}
  \city{Shenzhen}
  \country{China}
}
\email{kevinxwu@tencent.com}

\author{Tong Xu \dag}
\affiliation{%
  \institution{University of Science and Technology of China}
  \city{Hefei}
  \country{China}
}
\email{tongxu@ustc.edu.cn}

\author{Wanyu Wang  }
\affiliation{%
  \institution{City University of Hong Kong}
  \city{Hong Kong}
  \country{China}
}
\email{wanyuwang4-c@my.cityu.edu.hk}

\author{Yuyang Ye  }
\affiliation{%
  \institution{City University of Hong Kong}
  \city{Hong Kong}
  \country{China}
}
\email{yuyangye@cityu.edu.hk}

\author{Xiangyu Zhao \dag }
\affiliation{%
  \institution{City University of Hong Kong}
  \city{Hong Kong}
  \country{China}
}
\email{xianzhao@cityu.edu.hk}

\author{Enhong Chen}
\affiliation{%
  \institution{University of Science and Technology of China}
  \city{Hefei}
  \country{China}
}
\email{cheneh@ustc.edu.cn}

\author{Yefeng Zheng}
\affiliation{%
  \institution{Medical Artificial Intelligence Lab, Westlake University \& Jarvis Research Center, Tencent YouTu Lab}
  \city{Shenzhen}
  \country{China}
}
\email{zhengyefeng@westlake.edu.cn \&}
\email{yefengzheng@tencent.com}

\renewcommand{\shortauthors}{Derong Xu et al.}

\begin{abstract}
Model editing aims to precisely alter the behaviors of large language models (LLMs) in relation to specific knowledge, while leaving unrelated knowledge intact. This approach has proven effective in addressing issues of hallucination and outdated information in LLMs. However, the potential of using model editing to modify knowledge in the medical field remains largely unexplored, even though resolving hallucination is a pressing need in this area.
 Our observations indicate that current methods face significant challenges in dealing with specialized and complex knowledge in medical domain.
Therefore, we propose MedLaSA, a novel \textbf{\underline{La}}yer-wise \textbf{\underline{S}}calable \textbf{\underline{A}}dapter strategy for medical model editing. MedLaSA harnesses the strengths of both adding extra parameters and locate-then-edit methods for medical model editing. 
We utilize causal tracing to identify the association of knowledge in neurons across different layers, and generate a corresponding scale set from the association value for each piece of knowledge. Subsequently, we incorporate scalable adapters into the dense layers of LLMs. These adapters are assigned scaling values based on the corresponding specific knowledge, which allows for the adjustment of the adapter's weight and rank. The more similar the content, the more consistent the scale between them. This ensures precise editing of semantically identical knowledge while avoiding impact on unrelated knowledge.
To evaluate the editing impact on the behaviours of LLMs, we propose two model editing studies for medical domain: (1) editing factual knowledge for medical specialization and (2) editing the explanatory ability for complex knowledge.  We build two novel medical benchmarking datasets and introduce a series of challenging and comprehensive metrics. Extensive experiments on medical LLMs demonstrate the editing efficiency of MedLaSA, without affecting unrelated knowledge.  

\end{abstract}

\ccsdesc[500]{Applied computing~Life and medical sciences}
\keywords{Large Language Model, Model Editing, Medical Sciences}


\maketitle

\section{Introduction}\label{sec:intro}
The emergence of large language models (LLMs) has greatly promoted the development of natural language processing due to their powerful text generation capabilities \cite{zhao2023survey, li2024contextual, chen2024autoprm, xu2023large}. Meanwhile, LLMs also garner increasing attention from researchers in the healthcare field \cite{thirunavukarasu2023large,he2023survey}.
Although LLMs have proven to be valuable tools, they may still provide outdated factual information or even experience hallucinations \cite{zhang2023siren,li2023halueval,ye2023cognitive}, which is particularly concerning when deployed in real-world medical scenarios \cite{grace,feng2023trends,liu2024moe}.
To solve this problem, considering the substantial cost of retraining an LLM, there has been an increasing interest in model editing (also known as knowledge editing) \cite{wang2023easyedit}. It seeks to modify the behaviors of LLMs by precisely manipulating a part of knowledge while ensuring unrelated knowledge is unaffected (not the target of editing) \cite{zhang2024comprehensive,rome,memit,wang2023easyedit}. Researches have demonstrated that the LLMs can serve as a knowledge base for storing factual information about the world \cite{petroni2019language, tang2023counterfactual, geva2022transformer} and the knowledge in LLMs can be edited by understanding mechanization of knowledge retention within the transformer structure \cite{kn,geva2021transformer,lv2024interpreting}.

\begin{figure}[!t]
\centering
\includegraphics[width=0.7\linewidth]{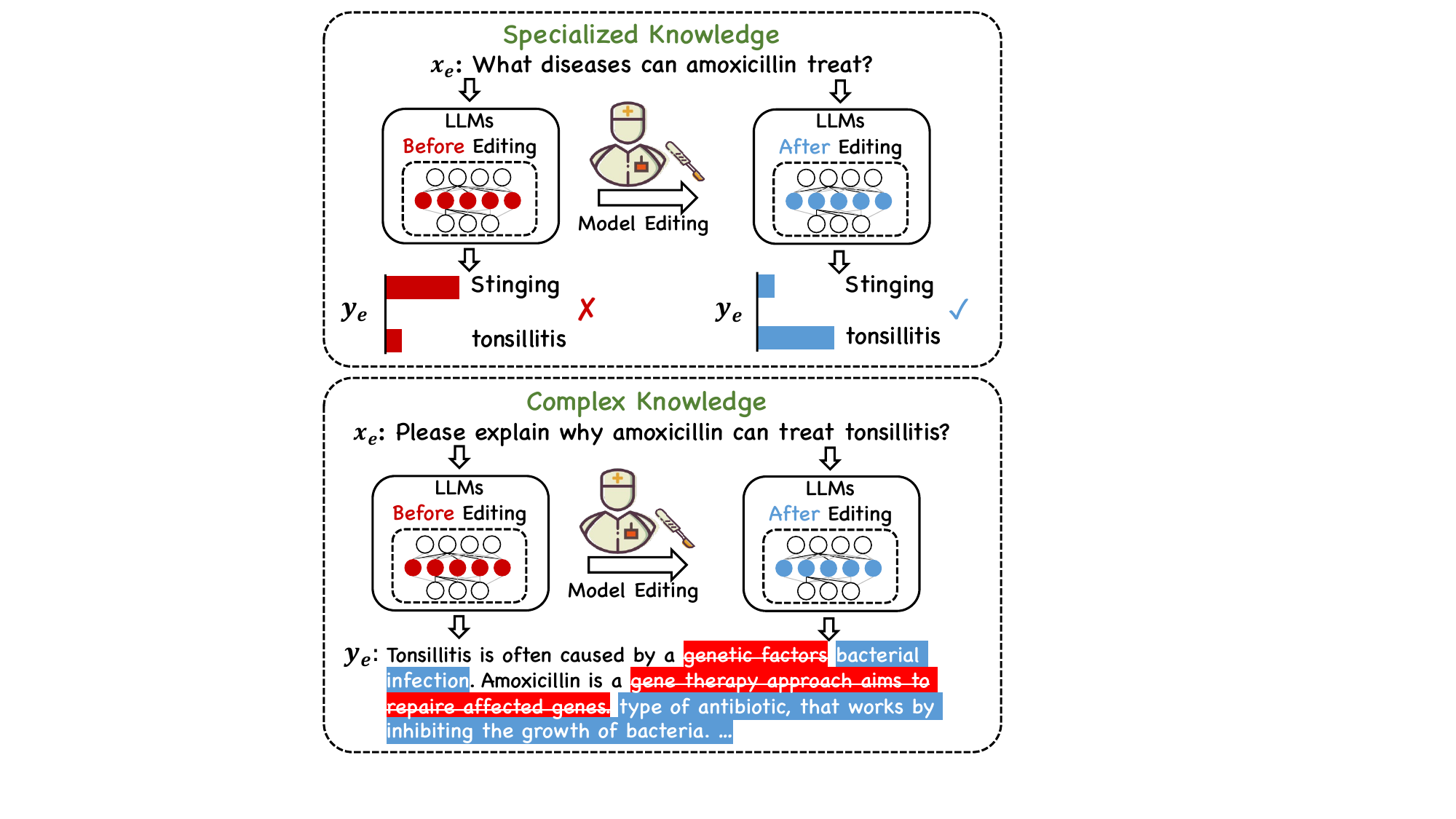}
\caption{Examples of specialized and complex knowledge in the medical domain. The red/blue symbols and background colors represent LLMs' state and output before/after model editing.
}
\label{fig:intro}
\end{figure}

Despite remarkable achievements of model editing methods in encyclopedic knowledge of the general domain \cite{rome,mend,memit}, the potential of utilizing model editing to modify knowledge in the medical field remains largely unexplored. 
They can be mainly categorized into two types: (i) Additional parameters or memories, aiming to introduce explicit memory to store editing knowledge \cite{serac,ike}, or add extra neurons to the network \cite{tpatcher}.  (ii) Modifying LLMs’ parameters, such as location-then-edit methods \cite{rome,memit}, which identify parameters corresponding to specific knowledge and modify them through direct updates to the target parameters.

We observed that these methods struggle to overcome the challenges of specialized and complex knowledge in the medical domain \cite{karabacak2023embracing}. As shown in Figure \ref{fig:intro}, LLMs may exhibit hallucinations and provide answers that do not align with the real world. It lacks specialized knowledge regarding the diseases that can be treated with amoxicillin \cite{chang2024survey,kung2023performance}, leading to incorrect responses such as "stinging". When dealing with complex knowledge, it is common to employ advanced prompting techniques (e.g., Chain-of-Thought \cite{lyu2023faithful}), to facilitate logical reasoning and deduce rational conclusions. However, LLMs may fail to explain the rationale behind the impact of amoxicillin in treating tonsillitis, resulting in an erroneous reasoning chain.
Current model editing methods either overlook the storage of complex medical knowledge across different layers in LLMs \cite{ike} or introduce substantial modifications to original parameters, which consequently affect the model's unrelated knowledge and lead to sub-optimal performance \cite{zhang2023siren}.
This research gap presents a question: \emph{Can model editing techniques effectively address challenges when integrating LLMs in the medical domain?}

To combine the strengths of both adding extra parameters \cite{tpatcher} and locate-then-edit methods \cite{rome}, we propose MedLaSA, a novel \textbf{\underline{La}}yer-wise \textbf{\underline{S}}calable \textbf{\underline{A}}dapter strategy specifically designed for medical model editing. MedLaSA first incorporates causal tracing, a knowledge localization tool, to establish connections between medical knowledge and corresponding layers. Then, for each piece of knowledge (represented as a QA pair), we construct an impact matrix that considers the influence of each token and layer on the knowledge. By aggregating the layer-wise impact matrix, we generate an impact set for each piece of knowledge. Instead of fine-tuning parameters \cite{rome}, we introduce adapters to dense weights.
We believe that by introducing additional parameters to edit the knowledge in high-influence layers (e.g., layer $i$), we can achieve a higher editing success rate while minimizing the impact on unrelated knowledge (such as knowledge with stronger connections to layer $j$).
It is important to note that our approach does not involve adding adapters to one layer while excluding others (we refer to this as hard scaling). Instead, our proposed method employs soft scaling, wherein we generate a corresponding scale set from the impact set for each knowledge. This allows for the adjustment of the adapter's weight and rank, ensuring the scalability of model editing. Each knowledge will have its own scale set, with the scale being more consistent for similar contents. Therefore, it can ensure accurate editing of the same knowledge while avoiding affecting unrelated knowledge.

On the other hand, there is a lack of benchmarks for editing medical knowledge. To address this gap and promote model editing techniques for mitigating hallucination issues in medical LLMs, we propose two initial studies, as illustrated in Figure \ref{fig:intro}: (1) Editing factual medical knowledge within LLMs to tackle the challenge of specialized knowledge, and (2) Editing LLMs to enhance their ability to effectively explain complex knowledge.
To achieve these, we construct two corresponding benchmarks, namely Medical Counter Fact (MedCF) and Medical Fact Explanation (MedFE). We propose to evaluate model editing from four perspectives: \textit{Efficacy}, \textit{Generality}, \textit{Locality}, and \textit{Fluency} \cite{zhang2024comprehensive}. Additionally, given the higher demand for reliability in the medical domain \cite{zhou2023survey}, we propose more comprehensive metrics to evaluate the post-edited model, addressing the following challenges:
\begin{itemize}[topsep=0pt, partopsep=0pt, leftmargin=15pt, parsep=0pt, itemsep=0pt]
    \item \textbf{Target Distribution}: Does the editing operation alter the probability distribution of the ground truth tokens?
    \item \textbf{Entity Mapping}: Does the editing operation solely learn the mapping relationship between head and tail entities?
    \item \textbf{Structural Similarity}: Does the editing operation influence unrelated knowledge with similar graph structures?
    \item \textbf{Textual Similarity}: Does the editing operation have an impact on unrelated knowledge that contains similar semantic text?
    \item \textbf{Consistent Topic}: Does the editing operation affect unrelated knowledge with the same topic?
\end{itemize}


Extensive experiments conducted on two comprehensive benchmarks demonstrate the superior performance of MedLaSA across a range of metrics, showing that it can achieve a higher editing success rate in specialized and complex medical knowledge contexts and avoid impacting irrelevant contexts. To the best of our knowledge, we are the first to propose model editing for enhancing factual knowledge and explanatory abilities in medical LLMs.\footnote{\url{https://github.com/quqxui/MedLaSA}}


\section{Related Work}\label{sec:relatedwork}
We present current model editing works following \citet{yao2023editing}.
\noindent
\textbf{Memories or Additional Parameters.}
The methods of with memories or additional parameters typically involve creating explicit memories to store the required knowledge for editing, or adding additional trainable parameters to LLMs for learning new knowledge \cite{melo,dong2022calibrating,grace,tang2022knowledge}.
SERAC \cite{serac} utilized explicit memory for storing edits and incorporated a scope classifier to understand the editing scope. Given a sample within the editing scope, it utilized a separate model to make edits, ensuring that the original model remains unaffected.
IKE \cite{ike} designed demonstration formatting and organization strategies, including the copy, update, and retain templates, and retrieved relevant knowledge facts from the editing memories as demonstration inputs to guide the editing process.
T-Patcher \cite{tpatcher} retained all original parameters to preserve overall performance while adding trainable neuron patches to the last Feed-Forward Network (FFN) layer of a Transformer for handling sequential model editing.
Despite their success, the above methods lack the exploration of the mechanics of knowledge storage in LLMs, which ultimately leads to poor performance in handling complex medical knowledge.
\noindent
\textbf{Modifying LLMs’ Parameters.}
The methods of this category aim to comprehend how knowledge is stored in LLMs and how it can be effectively altered by changing the parameters \cite{de2021editing,geva2021transformer,wu2023depn}.
KN \cite{kn} proposed a knowledge attribution method to identify the neurons associated with specific knowledge without fine-tuning, updating facts, and erasing relations by directly modifying the corresponding parameters in FFN.
MEND \cite{mend} introduced auxiliary hyper-networks to transform the gradient during the fine-tuning process, and trained the hyper-networks to ensure edit success and locality when updating LLMs' parameters.
ROME \cite{rome} applied Causal mediation analysis \cite{pearl2022direct,vig2020investigating} to identify decisive neuron activation and modify FFN weights by solving a least squares problem with a linear equality constraint using the Lagrange multiplier.
As an extension of ROME \cite{rome}, MEMIT \cite{memit} introduced a multi-layer algorithm to update multiple cases simultaneously.
Despite impressive progress made by these methods, they often introduce significant modifications to the original parameters. Consequently, unrelated knowledge is affected, resulting in a noticeable impact on \emph{Locality} and \emph{Fluency}, as demonstrated in Section \ref{sec:experiments} Experiments.

Overall, both of them have their advantages and disadvantages. Neuron localization can achieve interpretability between parameters and knowledge, but sometimes optimizing too much can have negative impacts on unrelated knowledge \cite{rome,geva2021transformer}. On the other hand, adding extra parameters has a smaller impact on unrelated knowledge but lacks exploration of the mechanics of knowledge storage \cite{tpatcher}.
Our proposed method, MedLaSA, aims to combine the strengths of both approaches. It identifies the association of knowledge in neurons of different layers, and introduces scalable adapters into the dense layers, enabling precise editing without affecting unrelated knowledge. 

\section{Methodology}\label{sec:method}

\subsection{Prelimimaries}\label{sec:prelim}
Model editing is a recently emerging field that aims to modify specific knowledge within a neural network while preserving the network's behaviours for other knowledge \cite{zhang2024comprehensive,yao2023editing}.
In contrast to vanilla fine-tuning for updating LLMs, model editing seeks to precisely manipulate and update the specific knowledge in LLMs, resulting in a more thorough and strict evaluation \cite{wang2023knowledge}. 
Formally, we denote a model as $f(x,\theta)$,  which maps an input $x$ to its prediction $y$ with the pre-trained model parameters $\theta$, and the post-edited model is denoted as $f'(x,\theta_e)$.
To be considered effective, model editing typically needs to satisfy the following four properties \cite{tpatcher,zhang2024comprehensive}:

\noindent
\textbf{Property 1 \textit{Efficacy}.}\quad The post-edited model should establish an effective mapping between the edit pair ($x_e$, $y_e$), i.e., $f'(x_e,\theta_e)=y_e$. The \textit{Efficacy} is calculated as the average accuracy on the edit case:
\begin{equation}
\mathbb{E}_{x'_e,y'_e} \mathbb{I} (argmax_{y} f'(y|x'_e,\theta_e)=y'_e).
\end{equation}

\noindent
\textbf{Property 2 \textit{Generality}.}\quad When an input sentence $x_g$ with a similar meaning to $x_e$ (e.g., a rephrased sentence) is provided, the post-edited model is expected to produce the corresponding output $y_e$ as well, i.e., $f'(x_g,\theta_e)=y_e$. The \textit{Generality} is computed as follows:
\begin{equation}
\mathbb{E}_{x'_g,y'_e} \mathbb{I} (argmax_yf'(y|x'_g,\theta_e)=y'_e).
\end{equation}

\noindent
\textbf{Property 3 \textit{Locality}.}\quad The editing process should remain local and precise, meaning the post-edited model should not impact the prediction of irrelevant pairs ($x_i$,$y_i$), i.e., $f'(x_i,\theta_e)=y_i$. Hence, the \textit{Locality} is evaluated by the rate at which the post-edited model $f'(x,\theta_e)$’s predictions are unchanged as the pre-edited model $f(x,\theta)$:
\begin{equation}
\mathbb{E}_{x'_i,y'_i} \mathbb{I} (f'(y|x'_i,\theta_e)=f(y|x'_i,\theta)).
\end{equation}

\noindent
\textbf{Property 4 \textit{Fluency}.}\quad The post-edited model should maintain generation ability and thus a high level of fluency in output. Following in \citet{rome}, \textit{Fluency} is calculated by measuring the weighted average of bi- and tri-gram entropies given by $-\sum_k{f_n(k)log_2f_n(k)}$, where $f_n(\cdot)$ is the n-gram frequency distribution, $k$ is output sentence of the model. 
 
\begin{figure*}[!t]
\centering
\includegraphics[width=0.9\linewidth]{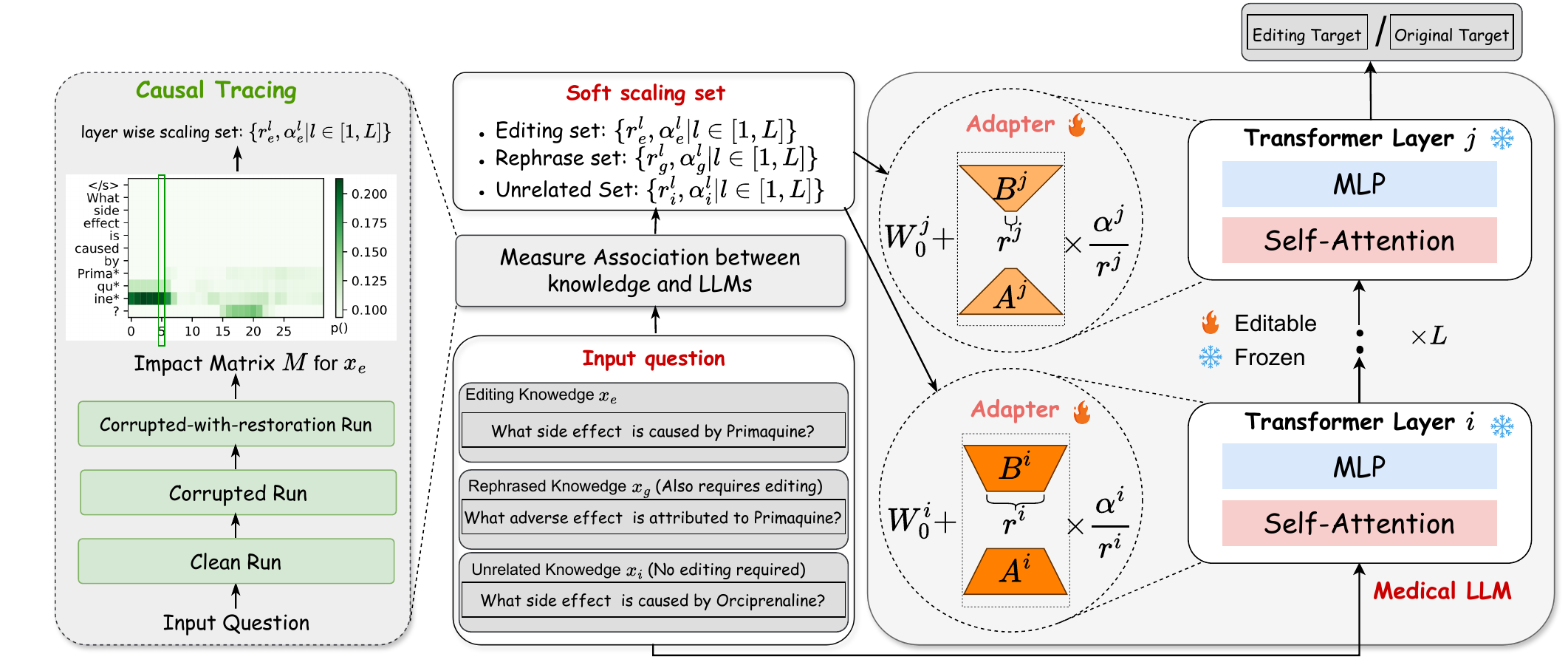}
\caption{The overview of our MedLaSA. We demonstrate the process of inputting editing, rephrased, and unrelated knowledge.
}
\label{fig:framework}
\end{figure*}
 
\subsection{Causal Tracing}\label{sec:Causal}
We first explain causal tracing, which aims to identify factual associations to specific neuron activation by calculating the contribution of each state towards factual predictions \cite{rome}. The knowledge and its associations in the network can be effectively utilized to regulate model editing and scaling operations in our model, as described in Section \ref{sec:medlasa}.
 This process involves three forward propagation runs: 
 
 \noindent
(1) \textbf{Clean run.} A question $x$ is fed into LLM, and the model predicts the probability of an answer $y$. For example, if the input is ``What diseases can amoxicillin treat?'', the output probability for ``tonsillitis'' is obtained. Meanwhile, the hidden activation value $\{h_i^l| i\in [1,T], l\in [1, L]\}$ of every token $i$ of $T$ tokens and every layer $l$ of $L$ layers are collected.

\noindent
(2) \textbf{Corrupted run.} The question $x$ is obfuscated by introducing Gaussian noise $\epsilon \sim N(0;v)$ with zero mean and standard deviation of $v$ to the subject embedding of $x$. Then, after going through the same forward propagation process as ``clean run'', the output probability for ``tonsillitis'' decreases due to the noise. Also, we can get a set of corrupted hidden activation $\{h_{i*}^l| i\in [1,T], l\in [1, L]\}$.

\noindent
(3) \textbf{Corrupted-with-restoration run.} Similar to the ``corrupted run'', the same input and noise are used. For each token within every layer, we separately replace the hidden activation value $h_{i*}^l$ with $h_{i}^l$ in the clean run. This restoration operation produces a probability of restoring the correct output for each token within every layer. A higher probability of token $i$ in layer $l$ indicates a stronger causal association between question and hidden states $h_{i}^l$. The probabilities can form an impact matrix $\boldsymbol{M} \in \mathbb{R}^{T \times L}$:
\begin{equation}
    \boldsymbol{M}_i^l = p(y|x,\epsilon, \theta_{h_{i}^l \leftarrow h_{i*}^l} ).
\end{equation}

The goal of causal tracing is to observe if the restoration of these hidden activations can recover the output probability of "tonsillitis" from the clear run. After iterating all the token-layer combinations, the token-layer combinations that successfully restore the output probability are considered highly relevant to the specific knowledge in the input "What diseases can amoxicillin treat?".
 
\subsection{MedLaSA}\label{sec:medlasa}

In this section, we introduce MedLaSA, a simple yet effective model editing strategy. MedLaSA is designed to modify each layer in a tailored manner by taking into account the associations between multiple layers and medical knowledge.
We first apply causal tracing to medical subject tokens (as described in Section \ref{sec:Causal}), which has been proven effective in identifying specific hidden states that are crucial when recalling a fact \cite{rome}.
Unlike previous methods such as ROME \cite{rome}, which directly modify the MLP weights of corresponding layers, we argue that adding an adapter to dense weights is a more effective way to insert new knowledge while mostly preserving the original abilities of LLMs.

Our motivation lies in that different knowledge will have different scale sets. Similar contents typically share more consistent scales between them. As a result, it ensures accurate editing of identical knowledge while avoiding impact on unrelated knowledge.
For instance, as illustrated in Figure \ref{fig:framework}, when data that needs editing is inputted, we increase its scale on the $i$-th layer and decrease its impact on the $j$-th layer, ensuring that knowledge updates occur on $i$-th layer. Conversely, when unrelated data is input, the influence of adapter on $j$-th layer is activated and reduces the impact of the adapter on $i$-th layer.
Inspired by LoRA \cite{lora}, we incorporate parallel low-rank adapter into dense layers to enable parameter-efficient fine-tuning. Specifically, a pre-trained weight matrix $\boldsymbol{W}_0\in \mathbb{R}^{d \times k}$ from attention or MLP module is updated with two trainable matrics $\boldsymbol{B}\in \mathbb{R}^{d \times r}$ and $\boldsymbol{A}\in \mathbb{R}^{r \times k}$. For layer $l$, we have:
\begin{equation}
    \boldsymbol{h}^l = \boldsymbol{W}_0^l\boldsymbol{x} + \frac{\alpha^l}{r^l}\boldsymbol{B}^l\boldsymbol{A}^l\boldsymbol{x},
\end{equation}
where, the rank $r^l \ll min(d,k)$ and the $\alpha^l$ in our work are utilized to adjust the number and location of trainable parameters required for knowledge updating, respectively.

\paragraph{\textbf{Scaling $\alpha$.}}
Parameter $\alpha$ is used to measure the weight of the adapter relative to the original network.
Each individual knowledge has a specific $\alpha$ value and can generate an impact matrix $\boldsymbol{M}$ through causal tracing in Section \ref{sec:Causal}. The impact $\boldsymbol{I}_\alpha$ is calculated as follows:
\begin{equation}
    \boldsymbol{I}_\alpha = norm(\sum_{t\in \mathbb{E}_s} \boldsymbol{M}_{t}),
\end{equation}
where $\mathbb{E}_s$ is defined as the set of the indexes of subject tokens, $norm(\cdot)$ denotes max-min normalization. The final scale of layer $l$ can be computed by multiplying hyper-parameter $\alpha_{o}$ with $\boldsymbol{I}_\alpha$ value in the $i$-th layer: $\alpha^l = \alpha_{o} \times I^l_\alpha$.

\paragraph{\textbf{Scaling Rank $r$.}}
The rank $r$ is used to control the number of additional parameters required to update new knowledge, which is generalized for knowledge in whole dataset $\mathbb{D}$ and is specific to each layer $l$:
\begin{equation}
    \boldsymbol{I}_r = norm(\sum_{M\in \mathbb{D}}\sum_{t\in \mathbb{E}_s} \boldsymbol{M}_{t}).
\end{equation}
The final rank of layer $l$ can be computed by multiplying hyper-parameter $r_{o}$: $r^l = \lceil r_{o} \times I^l_r\rceil$, which ensures the scalability of model editing to different knowledge.
By computing scale values $\alpha$ and $r$, we can create a set of layer-wise scaling factors \(\{r^l, \alpha^l| l\in [1, L]\}\) for editing, rephrased, and unrelated knowledge, as shown in Figure \ref{fig:framework}. Besides, a Transformer block typically includes self-attention and MLP modules, whose effects can be separately analyzed with causal tracing.
For example, to measure the impact matrix $\boldsymbol{M}_{attn}$ of the attention module, the MLP calculation is cut off and frozen in its corrupted run state, so that it is not affected by the insertion of a clean state.
We conducted a thorough analysis to examine the influence of various weights and hyper-parameters on model editing, in Section \ref{sec:experiments}.

\subsection{Medical Model Editing Benchmarks}\label{sec:dataconstruc}

\begin{table}[t]
\centering
\begin{tabular}{p{7.8cm}}
\toprule
Medical Counter Fact Dataset \\
\midrule
\midrule
\textbf{Original Triple}: (Primaquine, side effect, Nausea) \\
\textbf{Question}: What side effect is caused by Primaquine? \\
\textbf{Rephrase}: What adverse effect is attributed to Primaquine?\\
\textbf{Ground Truth}: Nausea \\
\textbf{Counterfactual Edit Target}: Stinging\\
\textbf{\textit{Efficacy} QA Pair}: (Question, Counterfactual Edit Target)\\
\textbf{\textit{Generality} QA Pair}: (Rephrase, Counterfactual Edit Target)\\
\bottomrule
\toprule
Medical Fact Explanation Dataset \\
\midrule
\midrule
\textbf{Fact}: In obesity which of the following hormone levels is decreased? Adiponectin.\\
\textbf{Rephrase}: In cases of obesity, which hormone experiences a decrease in levels? Adiponectin.\\
\textbf{Explanation / Edit Target}: Adiponectin is an abundant adipose-derived protein and enhances insulin sensitivity and lipid oxidation ...\\
\textbf{\textit{Efficacy} QA Pair}: (Fact, Explanation)\\
\textbf{\textit{Generality} QA Pair}: (Rephrase, Explanation)\\
\bottomrule
\end{tabular}
\caption{The edit examples in MedCF and MedFE datasets.}
\label{tab:datasample}
\end{table}

We aim to investigate the effectiveness of model editing techniques in the medical domain. However, a notable limitation is the lack of standardized benchmarks for editing medical knowledge.
On one hand, there are potential concerns regarding specialized medical knowledge stored in medical LLMs (e.g., outdated information and hallucination), which could result in errors in diagnosis or treatment recommendations \cite{zhou2023survey}.
On the other hand, real-world medical scenarios involve a high level of complex knowledge, which has led to an increased emphasis on the explanatory ability of LLMs, such as their ability to show a logical chain-of-thought during the decision-making process \cite{karabacak2023embracing}.
Therefore, we construct a Medical Counter Fact (MedCF) dataset for (1) editing factual medical knowledge and a Medical Fact Explanation (MedFE) dataset for (2) editing explanation ability of LLMs.

As mentioned in Section \ref{sec:prelim}, model editing methods generally need to satisfy the following four properties: \textit{Efficacy}, \textit{Generality}, \textit{Locality}, and \textit{Fluency}. Among them, \textit{Efficacy}, \textit{Generality}, and \textit{Locality} require the construction of extra data (i.e., QA pairs) while \textit{Fluency} does not require it and only requires evaluating the fluency of the output text. We explain how to construct the data for each property as follows.

\subsubsection{\textbf{Efficacy and Generality Data Construction}} \label{Efficacy and Generality Data Construction}
The QA pair of \textit{Efficacy} ($x_e$, $y_e$) is the piece of knowledge to be edited, and it is also used to measure the success rate of model editing. The QA pair of \textit{Generality} ($x_g$, $y_e$) is used to measure the success rate of editing for different expressions of $x_e$. Below, we explain the construction methods for MedCF and MedFE, separately.
\paragraph{\textbf{Medical Counter Fact Dataset.}}
To build the MedCF dataset, we utilize a medical knowledge graph called DRKG  \cite{drkg2020,10231041}. For each triple $(h, r, t)$, we convert $h$ and $r$ into a question and $t$ into an answer, thus creating a QA pair. To evaluate the ability to edit knowledge with unknown prediction results of LLMs, we introduce counterfactual data for model training following ROME \cite{rome}. We randomly sample a negative entity $t_*$ from the knowledge graph and use it as a new answer, attempting to edit the original answer accordingly. For example in Table \ref{tab:datasample}, the original triple  ``(Primaquine, side effect, Nausea)'' is converted to the question: ``What side effect is caused by Primaquine?'' and the answer ``Nausea''. But the editing target is indeed a counterfactual result ``Stinging''.

In terms of the \textit{Generality} data (`Rephrase' text in Table \ref{tab:datasample}), we employ ChatGPT to rephrase the question. The design of the prompts can be found in Table \ref{tab:Templates}. For the MedCF dataset, we generated a question template and a rephrased template for each relation and applied them to all the triples. For instance, the relation template (\textit{head entity, side effect, tail entity}) is transformed to: ``What side effect is caused by [head entity]?'' This template-based QA generation can reduce the need for a large number of ChatGPT accesses and maintain the reliability of generated content quality.

\paragraph{\textbf{Medical Fact Explanation Dataset.}}
We build the MedFE dataset by utilizing MedMCQA \cite{pmlr-v174-pal22a}, a dataset designed for answering medical entrance exam questions. To generate `Fact' as shown in Table \ref{tab:datasample}, we combined the question and correct choice in MedMCQA to form a factual statement, and we used the expert's explanation in MedMCQA as a source for the `Explanation / Edit Target'.

The data of \textit{Efficacy} for MedCF and \textit{Generality} for both MedCF and MedFE are generated by querying ChatGPT. 
The template for querying is shown in Table \ref{tab:Templates}. This prompt template ensures that the generated content adheres to medical terminology while preserving the original meaning of the question.
Different from MedCF, all the facts require paraphrasing for the MedFE dataset.

To ensure the credibility and quality of the text generated by ChatGPT, we conducted additional checks. For the MedCF, ChatGPT was employed to convert triplets into questions and rephrase text. As explained in Section \ref{Efficacy and Generality Data Construction}, this process involved transforming relation templates, resulting in the need to verify only 34 data points. In the case of the MedFE, questions were already known, and ChatGPT was solely used for rephrasing. In total, there were 4225 data points. We thoroughly examined all the generated data and took measures to ensure that the rephrased sentences maintained their original meaning.
\begin{table}[t]
\centering
\begin{tabular}{p{7.8cm}}
\toprule
Templates for Querying LLMs\\
\midrule
\midrule
\textbf{\textit{Efficacy}}: Given a triplet ([$h$], [$r$], [$t$]), please express this triplet in a question-answer form. The [$h$] and [$r$] form a question, and you need to ask what the corresponding answer [$t$] is for this question. \\
\textbf{\textit{Generality}}: Please rephrase the question using medical terminology, without changing the semantics: [$q$]\\
\bottomrule
\end{tabular}
\caption{Templates for querying ChatGPT (gpt-3.5-turbo-0613) to generate the \textit{Efficacy} and \textit{Generality} data. $h$, $r$, and $t$ denote the head entity, relation, and tail entity, respectively.}
\label{tab:Templates}
\end{table}

\subsubsection{\textbf{Locality Data Construction.}}\label{sec:ConstructingLocalityEvaluation}
The QA pair of \textit{Locality} ($x_i$,$y_i$) aims to assess the impact of model editing on unrelated knowledge. Previous benchmarks either employed out-of-distribution data (e.g., zsRE \cite{mend}) or solely relied on data with the same ground truth (e.g., CounterFact \cite{rome}). Nevertheless, we argue that a comprehensive evaluation of \textit{Locality} is necessary to prevent the inadvertent modification of irrelevant knowledge and ensure high reliability of the medical domain.
As we explained in Section \ref{sec:intro}, the post-edited model should be evaluated based on the following challenges: Target Distribution (TD), Entity Mapping (EM), Structural Similarity (SS), Textual Similarity (TS), and Consistent Topic (CT). The issues that these evaluation metrics aim to address are mentioned in Section \ref{sec:intro}.
To fulfill the requirements, we gathered relevant QA pairs for evaluation.
The original triple of \textit{Efficacy} QA pair is denoted as $(h, r, t)$. We employ the graph structure of KG for searching and building \textit{Locality} sub-metrics. For \textit{Locality}-TD, we sample new triples $(h_{td}, r_{td}, t)$ by fixing $t$ to investigate the shift of the ground-truth token. For \textit{Locality}-EM, we sample new triples $(h, r_em, t_em)$ by fixing $h$ to create mappings between head and tail entities. For \textit{Locality}-SS, we utilize the knowledge graph embedding method, RotatE \cite{sun2018rotate}, to learn the entity and relation embeddings from the graph structure. For \textit{Locality}-TD, we use BioBERT \cite{lee2020biobert} to convert $(h, r, t)$ triples into text embeddings, and then obtain similar triples $(h_s, r_s, t_s)$ based on Structural Similarity and Textual Similarity, respectively. For \textit{Locality}-CT, we utilize the source data from MedMCQA, which includes topic tags for each question, annotated by experts.
These new triples are also converted into QA pairs, which enable a thorough analysis of the impacts of model editing techniques on unrelated knowledge in medical domain. It should be noted that the calculation for all sub-metrics of \textit{Locality} is the same, as listed in Section \ref{sec:prelim}.

\begin{figure}[!t]
\centering
\includegraphics[width=0.45\linewidth]{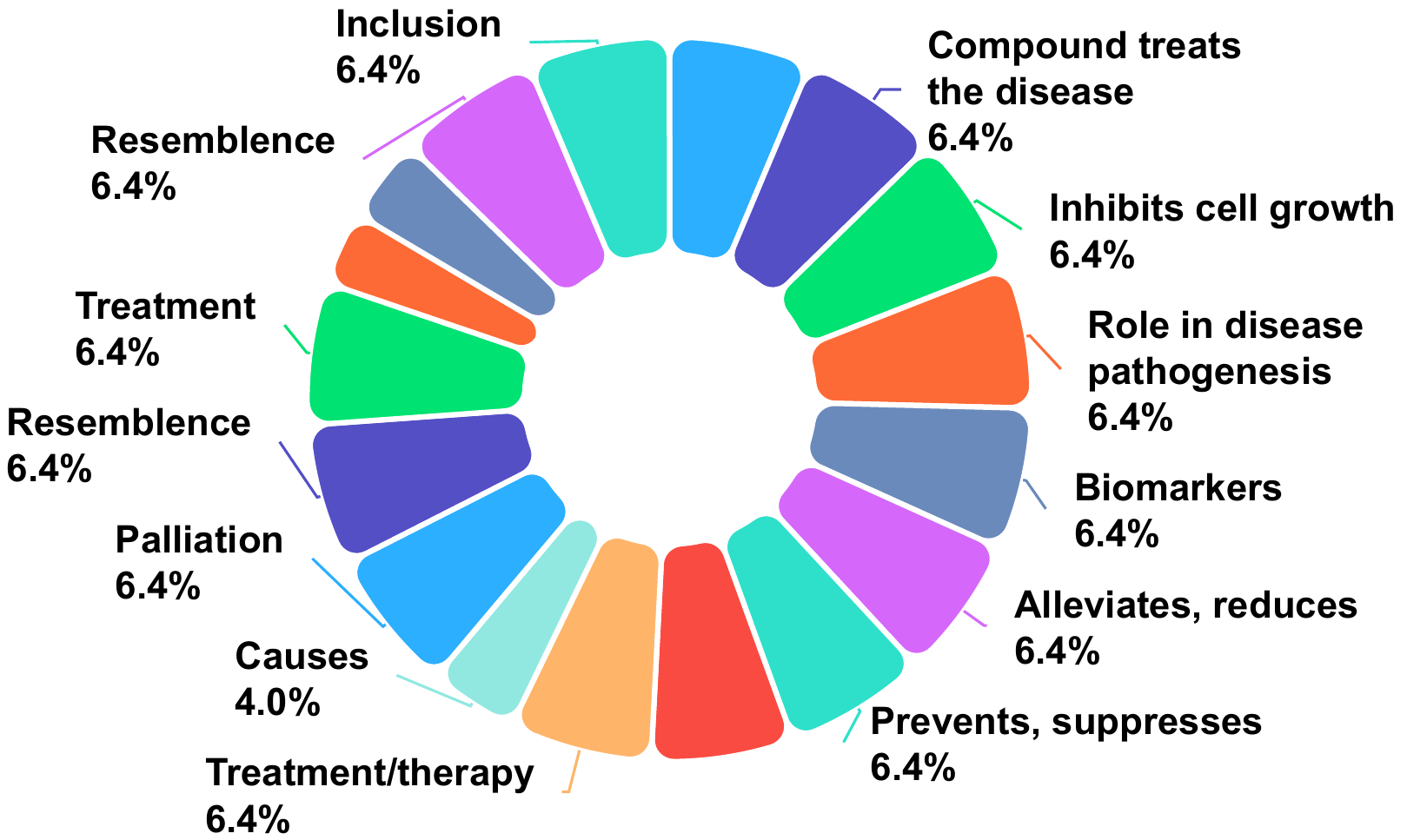}
\includegraphics[width=0.5\linewidth]{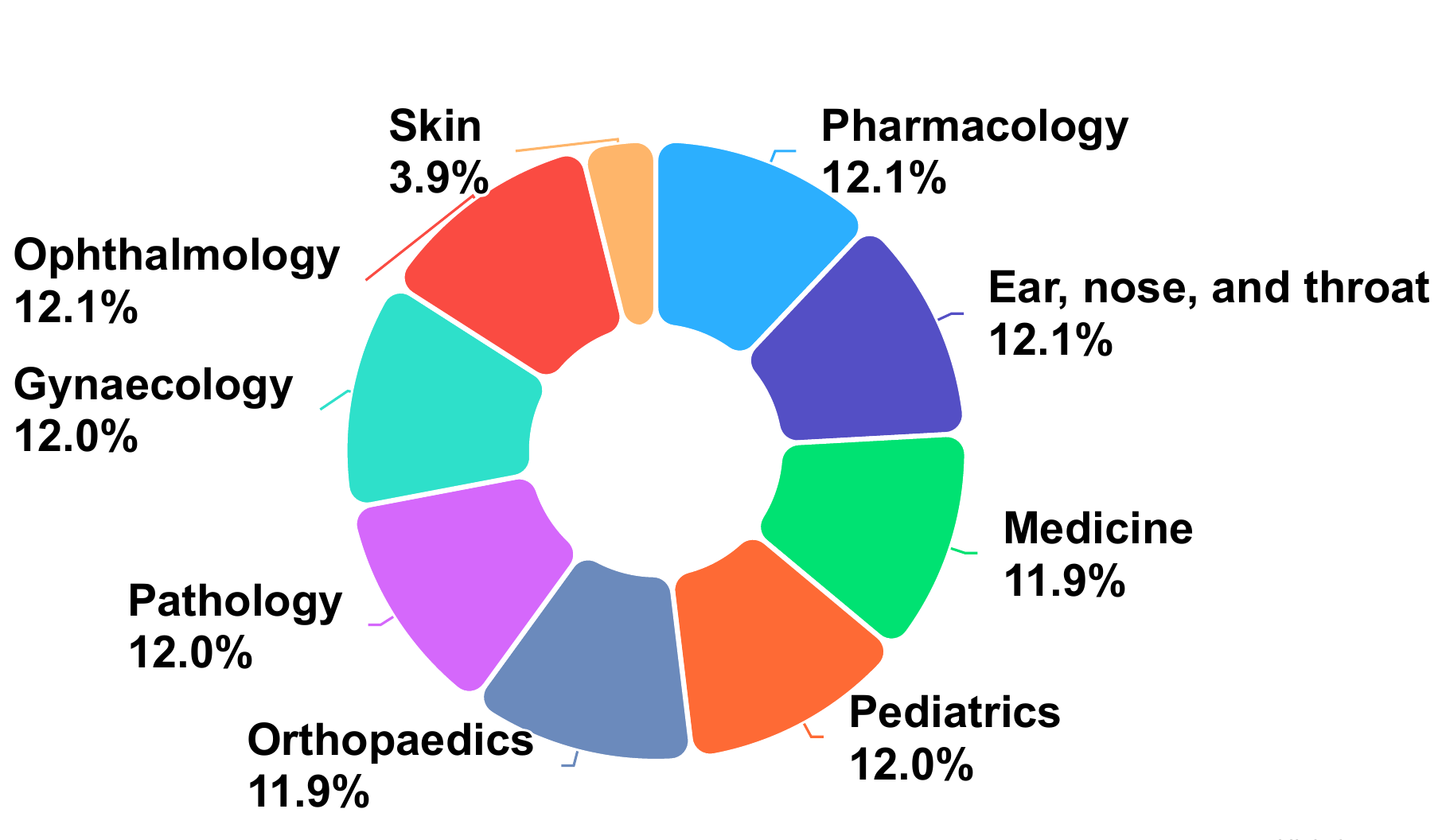}
\caption{Knowledge type distribution of MedCF (Left) and MedFE (Right) dataset.}
\label{fig:type}
\end{figure}

We provide knowledge type distribution of MedCF and MedFE in Figure \ref{fig:type}, respectively. The knowledge types in MedCF are categorized based on their relations, while the types in MedFE are categorized according to medical subjects. Data splits of MedCF and MedFE are shown in Table \ref{tab:Statistics}. We have attempted to ensure that different categories have similar proportions, but due to data limitations, some types have not been collected as extensively as others. More details about datasets can be found in our 
 repository.





\section{Experiments}\label{sec:experiments}


In this section, we present extensive experimental results to investigate the following research questions:
\begin{itemize}[topsep=0pt, partopsep=0pt, leftmargin=13pt, parsep=0pt, itemsep=0pt]
    \item \textbf{RQ1.} How does the proposed MedLaSA model perform compared to the state-of-the-art methods?
    \item \textbf{RQ2.} What impact do different layer-wise selection strategies have on performance?
    \item \textbf{RQ3.} How do the proposed modules improve the performance?
    \item \textbf{RQ4.} How do the settings of hyper-parameters affect the model editing performance?
    \item \textbf{RQ5.} What impact do different editable weights have on the model editing performance?
    \item \textbf{RQ6.} Does MedLaSA have the ability to learn the association between knowledge and parameters?
\end{itemize}

\subsection{Experimental Setup}

\paragraph{\textbf{Metrics.}}
we utilize the metrics constructed in Section \ref{sec:dataconstruc} in our evaluation. The computation of metrics follows EasyEdit \cite{wang2023easyedit}, which are measured as average accuracy between the token matching of the predicted output and the expected output.
For ease of presentation, we employ abbreviations to represent each metric: \textit{Efficacy} (Eff.), \textit{Generality} (Gen.), \textit{Locality} (Loc.), and \textit{Fluency} (Flu.). Due to limitations of original data, we measure TD, EM, SS, TS for MedCF and measure TS and CT for MedFE.
To examine the trade-off between edit success and locality, we further report the weighted mean by: 

\begin{equation}
    Average = ( \frac{Eff. + Gen.}{2} + \frac{\sum_{m\in Loc.} m}{|Loc.|}) /2.
\end{equation}

We use Average (Avg.) because the indicators of successful editing (\textit{Efficacy}, \textit{Generality}) and affecting unrelated knowledge (\textit{Locality}) can sometimes conflict with each other.
For instance, a model may achieve a high efficacy score by over-fitting the editing data, but this could result in a significant decrease in performance on unrelated knowledge. Therefore, we believe that a model editing method should keep the tread-off between successful editing and affecting unrelated knowledge.

\paragraph{\textbf{Backbones and Baselines.}}
As there is a lack of datasets for medical model editing, all our experiments are conducted on MedCF and MedFE datasets, using four NVIDIA V100 32G.
In our experiments, we specifically address the single-edit problem, which has also been adopted in previous studies \cite{rome,kn,tpatcher}. To tackle this problem, we utilize two medical LLMs, namely ChatDoctor-7B \cite{yunxiang2023chatdoctor} and Meditron-7B \cite{chen2023meditron70b}, which serve as the to-be-edited models in our research. We compare various model editing methods, including FT (Fine-tuning on multiple layers), LoRA \cite{lora}, ROME \cite{rome}, MEND \cite{mend}, and MEMIT \cite{memit}. The implementation of all baseline codes is based on the open-source tool EasyEdit \cite{wang2023easyedit}. We maintain same settings across all methods to ensure consistency. All hyper-parameters of baselines are set according to optimal values in the validation set of corresponding works. 
During the hyper-parameters selection of MedLaSA, we search for $r$, $\alpha$, and editable weights $W$. The search scope for $r$ and $\alpha$ was limited to the values [2, 8, 24, 32, 64, 128]. The analysis of $r$ and $\alpha$ can be seen in Figure \ref{fig:para}. The weights that were editable in ChatDoctor and Meditron include Attn weights $W_q$, $W_v$, $W_k$, $W_o$, and MLP weights $W_{up}$, $W_{down}$, $W_{gate}$, which is analysed in Table \ref{tab:weights}. We selected parameter with the highest average value in the validation set. 

\begin{table}[]
\centering
            \fontsize{9}{11}\selectfont
\setlength{\tabcolsep}{4.2mm}{
\begin{tabular}{l|rrrr}
\toprule
Dataset &\#Type	&\#Train &\#Valid &\#Test   \\
\hline\hline
MedCF 	&17 &2,407 & 817 & 801 \\
MedFE 	& 9 & 2,533 & 851 & 841 \\
\bottomrule
\end{tabular}}
\caption{Statistics of MedCF and MedFE.}
\label{tab:Statistics}
\end{table}

\subsection{Main Results (RQ1)}

\begin{table*}[t]

\centering
            \fontsize{9}{11}\selectfont
\setlength{\tabcolsep}{2mm}{
\begin{tabular}{l |  cccccc |  cccccccc}
\toprule
\multirow{3}{*}{Models} & \multicolumn{6}{c|}{MedFE}  & \multicolumn{8}{c}{MedCF}  \\
\cline{2-7}\cline{8-15}
 & \multirow{2}{*}{Eff.} & \multicolumn{1}{c}{\multirow{2}{*}{Gen.}}   & 
\multicolumn{2}{c}{Loc.} & \multicolumn{1}{c}{\multirow{2}{*}{Flu.}} & \multicolumn{1}{c|}{\multirow{2}{*}{Avg.}} & \multicolumn{1}{c}{\multirow{2}{*}{Eff.}}  & \multicolumn{1}{c}{\multirow{2}{*}{Gen.}}   & \multicolumn{4}{c}{Loc.} & 
\multicolumn{1}{c}{\multirow{2}{*}{Flu.}}  & \multicolumn{1}{c}{\multirow{2}{*}{Avg.}} \\
\cline{4-5}\cline{10-13}
 \multicolumn{1}{c|}{}  & \multicolumn{1}{c}{} & \multicolumn{1}{c}{}   &CT&TS&\multicolumn{1}{c}{}   & \multicolumn{1}{c|}{} &  \multicolumn{1}{c}{}  & \multicolumn{1}{c}{}   & TD&EM&SS&TS& \multicolumn{1}{c}{}\\ 
\hline
\hline
\multicolumn{15}{c}{ChatDoctor-7B} \\
\hline
FT &61.39&61.04&73.09&70.78&516.44&66.57 & 61.55&61.48&60.74&63.02&59.66&58.74&356.31&61.03   \\
LoRA & 94.45&88.56&83.24&79.75&570.12&86.50 & 72.01&71.90&93.52&91.88&91.76&92.72&575.71&82.21 \\
MEND &40.66&40.51&50.43&44.52&385.75&44.03 & 24.72&24.71&75.29&75.17&74.85&75.24&449.94&49.92\\
ROME &84.01&69.37&92.88&81.98&572.82&82.06 & 72.73&72.51&92.27&61.20&89.41&86.51&556.11&77.48\\
MEMIT &84.59&70.23&\textbf{95.80}&82.46&566.95&83.27& \textbf{82.20}&\textbf{82.03}&94.61&62.12&92.09&91.01&563.31&83.54\\
\rowcolor{gray!10} MedLaSA&\textbf{98.11}&\textbf{93.58}&89.25&\textbf{84.11}&\textbf{576.13}&\textbf{91.26}& 72.37&70.80&\textbf{96.16}&\textbf{95.24}&\textbf{95.59}&\textbf{95.19}&\textbf{583.49}&\textbf{83.56}\\
\hline
\multicolumn{15}{c}{Meditron-7B} \\
\hline
FT &62.82& 62.68&67.62& 64.94&473.66& 64.51&65.97& 65.36 &48.91& 50.39& 48.13& 46.25&  327.76& 57.04 \\
LoRA&94.01&89.29&83.75&79.13&571.42&86.55 & 72.19&71.80&92.29&91.11&91.36&92.42&572.33&81.90\\
MEND&34.21&31.34&30.03&34.23&404.19& 32.46 & 22.87&22.93&71.16&71.21&71.03&72.29&428.38&47.16 \\
ROME&84.59&69.22&\textbf{95.78}&\textbf{86.44}&564.75&84.01 & 72.69&72.91&92.79&61.80&90.06&86.93&559.82&77.84\\
MEMIT&84.91&70.80&95.40&82.02&566.95&83.28 & \textbf{83.10}&\textbf{83.23}&95.01&62.62&92.99&90.50&563.31&\textbf{84.22}\\
\rowcolor{gray!10} MedLaSA&\textbf{98.77}&\textbf{94.81}&87.41&81.67&\textbf{575.58}&\textbf{90.66} & 72.37&71.06&\textbf{95.71}&\textbf{94.84}&\textbf{95.04}&\textbf{94.90}&\textbf{582.80}&83.42\\
\bottomrule
\end{tabular}}
\caption{Model editing results compared with other state-of-the-art methods on MedCF and MedFE benchmarks. The best results are highlighted in bold, and larger values for all metrics indicate better performance. It should be noted that Locality metrics TD, EM, and SS necessitate source data that is structured in a knowledge graph format, and thus, can only be utilized for MedCF. Metric CT requires a more specific topic for the question, making it applicable only to MedFE.}
\label{tab:mainresults}
\end{table*}

We present the main results compared with baselines. As indicated in Table \ref{tab:mainresults}, MedLaSA demonstrates significant improvements over baselines across most metrics.
For instance, MedLaSA exhibits superior performance of \textit{Fluency} on both datasets and two LLM backbones, validating our method's ability to maintain generation capability.
The experimental results of FT indicate that excessive retraining of parameters could result in model collapse, causing the model to lose its original generating capability (i.e., much lower \textit{Fluency}).
LoRA introduces supplementary parameters but fails to consider the impact on unrelated knowledge. It also overlooks the specific positioning of knowledge in LLMs and dependency of knowledge on different layers. As a result, LoRA performs poorly in terms of \textit{Locality} compared to MedLaSA.
MEND has high requirements for initialization conditions, resulting in lower average performance on these medical datasets.
ROME focuses solely on single-layer knowledge editing of LLMs, without taking into account the knowledge stored in multi-layers, thus the performance tends to deteriorate.

MEMIT performs well in terms of \textit{Efficacy} and \textit{Generality} on MedCF because it is specifically designed to handle counterfactual data based on triplets, similar to MedCF. It edits counterfactual data by locating the key through the subject in the prompt and optimizing the value to select the object. This phenomenon is also evident in the case of ROME, as it has enabled ROME to achieve comparable or even superior performance to MedLaSA.
However, MEMIT and ROME fall short in terms of \textit{Locality} when compared to MedLaSA, particularly in Entity Mapping (EM value of MEMIT decreased by 33.12\% on ChatDoctor). This is because MEMIT and ROME only learn the mapping relationship between the head and tail entities, leading to consistent predictions when the subject in the locality prompt is the same as the editing prompt.
Furthermore, MEMIT's performance on the MedFE dataset is inferior due to its inability to handle long text output and complex multiple knowledge.
Due to its reliance on subject-to-object localization, MEMIT may not be suitable for scenarios that involve complex knowledge and longer text.
In contrast, our proposed MedLaSA addresses this issue by dynamically adjusting the scale of additional parameters and ensuring the insertion of complex knowledge.

\subsection{Strategies for Layer Selection (RQ2)}
In this section, we evaluate different strategies for layer selection, including the \underline{Random} and \underline{Fixed} strategies, in comparison to our layer-wise scalable adapter strategy, as shown in Table \ref{tab:strategies}. 
The \underline{Random} strategy involves randomly selecting the scale values of rank $r_{o}$ and alpha $\alpha_{o}$ of all layers, instead of using Causal tracing to determine knowledge location. The reported results were obtained through five random sampling experiments. The \underline{Fixed} strategy maintains fixed scale values of rank $r_{o}$ and alpha $\alpha_{o}$ to all data (factual knowledge), same with MedLaSA, across all layers.
From Table \ref{tab:strategies}, it is evident that our designed strategy outperforms both the \underline{Random} and \underline{Fixed} strategies across all metrics, which proves the effectiveness of MedLaSA.
Moreover, \underline{Random} strategy's performance is hindered by the unpredictability of parameter selection. This randomness leads to lower \textit{Efficacy} and lower \textit{Generality} compared to \underline{Fixed} strategy. Despite these shortcomings, the \underline{Random} strategy's varying scales for different knowledge and layers result in a lesser impact on irrelevant knowledge compared to the \underline{Fixed} strategy, leading to higher scores in terms of \textit{Locality} and \textit{Fluency}.

\begin{table}[t]

\centering
\fontsize{9}{11}\selectfont
\setlength{\tabcolsep}{2.1mm}{
\begin{tabular}{l|cccccc}
\toprule
\multicolumn{1}{l|}{\multirow{2}{*}{Strategy}} & \multicolumn{1}{c}{\multirow{2}{*}{Eff.}}  & \multicolumn{1}{c}{\multirow{2}{*}{Gen.}}   & \multicolumn{2}{c}{Loc.} & \multicolumn{1}{c}{\multirow{2}{*}{Flu.}} & \multicolumn{1}{c}{\multirow{2}{*}{Avg.}}\\
\cline{4-5}

 \multicolumn{1}{c|}{}    &  \multicolumn{1}{c}{}  & \multicolumn{1}{c}{}  &CT&TS&\multicolumn{1}{c}{} &\multicolumn{1}{c}{} \\ 
\hline\hline
Random& 92.15&86.98&85.62&80.11&572.76&86.22\\
Fixed & 94.45&88.56&83.24&79.75&570.12&86.50\\
\rowcolor{gray!10} MedLaSA & 98.77&94.81&87.41&81.67&575.58&90.66\\
\bottomrule
\end{tabular}}
 \caption{Comparison of different editing strategies on MedFE. }
\label{tab:strategies}
\end{table}

\subsection{Ablation Study (RQ3)}
In this section, we analyze the effects after removing Scaling $\alpha$ (SA) and Scaling Rank (SR) in the self-attention (Attn) and MLP layers, as shown in Table \ref{tab:ablation}.
We observe that removing SR leads to a decline in all metrics. This suggests that SR plays a crucial role in maintaining the overall performance, especially on the \textit{Locality} metrics.
On the other hand, when SA is removed, there is an improvement in the \textit{Generality}. This improvement, however, comes at the cost of significant decreases in \textit{Locality}-CT (from 89.25\% to 82.16\%) and \textit{Locality}-TS (from 84.11\% to 75.84\%), indicating that while SA helps in minimizing the model's modification of irrelevant knowledge, it concurrently compromises the model's generalization to rephrase.
When both SA and SR are removed, the overall performance declines the most. It is worth noting that there are significant decreases in the \textit{Efficacy} and \textit{Generality} metrics, showing a decline in the success rate of medical model editing.
Similar results can also be observed when our proposed method is applied exclusively to Attn weights or MLP weights, which further demonstrates the effectiveness of the proposed parameters SA and SR in medical model editing.

\begin{table}[t]
\centering
            \fontsize{9}{11}\selectfont
\setlength{\tabcolsep}{1.9mm}{
\begin{tabular}{l|cccccc}
\toprule
\multicolumn{1}{l|}{\multirow{2}{*}{Variant}} & \multicolumn{1}{c}{\multirow{2}{*}{Eff.}}  & \multicolumn{1}{c}{\multirow{2}{*}{Gen.}}   & \multicolumn{2}{c}{Loc.} & \multicolumn{1}{c}{\multirow{2}{*}{Flu.}} & \multicolumn{1}{l}{\multirow{2}{*}{Avg.}}\\
\cline{4-5}
 \multicolumn{1}{c|}{}    &  \multicolumn{1}{c}{}  & \multicolumn{1}{c}{}  &CT&TS&\multicolumn{1}{c}{} &\multicolumn{1}{c}{} \\ 
\hline\hline
\rowcolor{gray!10} ALL & 98.11&93.58&89.25&84.11&576.13&91.26 \\
w/o SR & 96.85&93.24&84.88&79.02&573.56&88.50\\
w/o SA & 96.90&94.44&82.16&75.84& 571.36&87.34\\
w/o SA\&SR & 94.45&88.56&83.24&79.75&570.12&86.50\\
\hline
w/o Attn & 97.41& 93.04&88.47& 83.82&573.34&90.68\\
w/o Attn\&SR & 96.11&91.57&87.19&82.10&574.85&89.24\\
w/o Attn\&SA & 96.86&94.12&87.51&80.19& 575.47&89.67\\
\hline
w/o MLP & 92.94&85.11&88.18&84.17& 578.41&87.60\\
w/o MLP\&SR & 90.96&84.13&86.19&82.20& 577.30&85.87\\
w/o MLP\&SA & 96.05&91.07& 84.57&79.27&576.13&87.74\\
\bottomrule
\end{tabular}}
\caption{Ablation study on MedFE. 
}
\label{tab:ablation}
\end{table}

\subsection{Hyper-parameters Analysis (RQ4)}

\begin{table*}[t]
\centering
            \fontsize{9}{11}\selectfont
\setlength{\tabcolsep}{3.8mm}{
\begin{tabular}{c|l|cccccc}
\toprule
\multicolumn{1}{c|}{\multirow{2}{*}{Weight Type}}&\multicolumn{1}{c|}{\multirow{2}{*}{Editable Weight}} & \multicolumn{1}{c}{\multirow{2}{*}{Eff.}}  & \multicolumn{1}{c}{\multirow{2}{*}{Gen.}}   & \multicolumn{2}{c}{Loc.} & \multicolumn{1}{c}{\multirow{2}{*}{Flu.}} & \multicolumn{1}{c}{\multirow{2}{*}{Avg.}}\\
\cline{5-6}
& \multicolumn{1}{c|}{}    &  \multicolumn{1}{c}{}  & \multicolumn{1}{c}{}  &CT&TS&\multicolumn{1}{c}{} &\multicolumn{1}{c}{} \\ 
\hline\hline
\multicolumn{1}{c|}{\multirow{7}{*}{Attn Weights}} & $W_q$ & 61.16 & 58.24 & 98.48 & 97.82 & 580.67 & 78.93 \\
& $W_v$ & 77.11 & 70.76 & 92.50 & 89.84 & 579.75 & 82.55 \\
& $W_k$ & 59.89 & 57.27 & 98.72 & 98.25 & 581.17 & 78.53 \\
& $W_o$ & 75.81 & 69.55 & 95.02 & 92.16 & 580.62 & 83.13 \\
& $W_q$, $W_v$, $W_k$, $W_o$ & 92.07 & 84.16 & 90.23 & 86.21 & 577.86 & 88.17 \\
\hline
\multicolumn{1}{c|}{\multirow{5}{*}{MLP Weights}} & $W_{up}$ & 82.62 & 76.30 & 96.47 & 93.40 & 580.81 & 87.20 \\
& $W_{down}$ & 80.73 & 74.88 & 94.09 & 91.03 & 578.29 & 85.18 \\
& $W_{gate}$ & 80.78 & 74.73 & 97.38 & 94.57 & 580.69 & 86.86 \\
& $W_{up}$, $W_{down}$, $W_{gate}$ & 96.47 & 90.93 & 91.89 & 87.31 & 576.53 & 91.65 \\
\hline
\multicolumn{1}{c|}{\multirow{3}{*}{Attn + MLP Weights}} & $W_q$, $W_v$, $W_{up}$, $W_{down}$ & 96.99 & 91.43 & 88.93 & 84.22 & 577.0 & 90.39 \\
& $W_q$, $W_v$, $W_{up}$, $W_{down}$, $W_{gate}$ & 98.27 & 93.70 & 88.73 & 83.58 & 576.13 & 91.07 \\
& $W_q$, $W_v$, $W_k$, $W_o$, $W_{up}$, $W_{down}$, $W_{gate}$ & 98.70 & 94.63 & 87.66 & 82.10 & 574.19 & 90.77 \\
\bottomrule
\end{tabular}}
\caption{Comparison of the impact of different editable weights in ChatDoctor, which is based on Llama \cite{touvron2023llama} and includes Attn weights ($W_q, W_v, W_k, W_o$) and MLP weights ($W_{up}, W_{down}, W_{gate}$).}
\label{tab:weights}
\end{table*}

\begin{figure}[t]
\centering
\includegraphics[width=0.4\columnwidth]{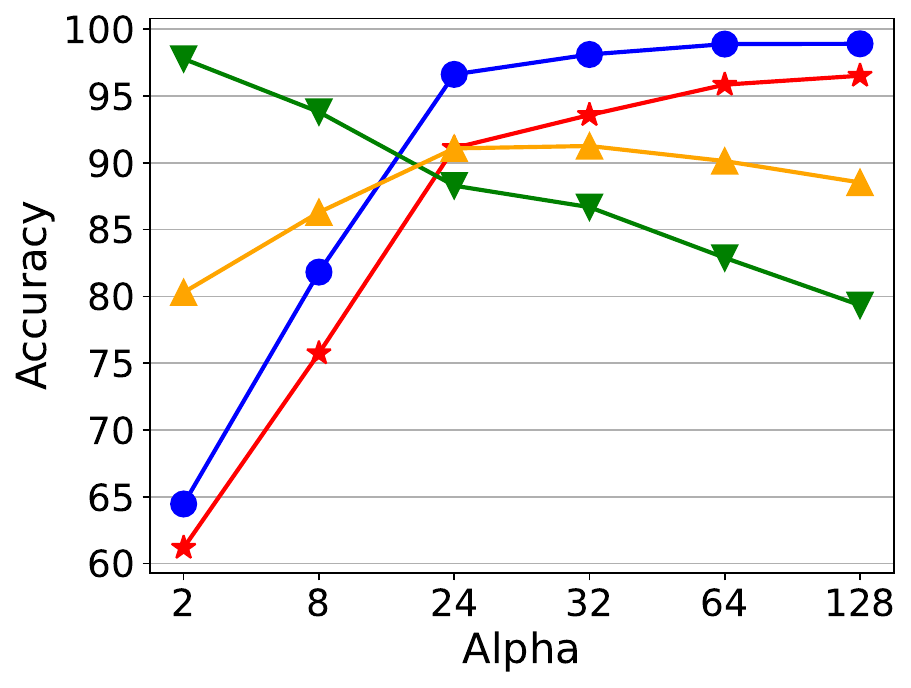}
\includegraphics[width=0.4\columnwidth]{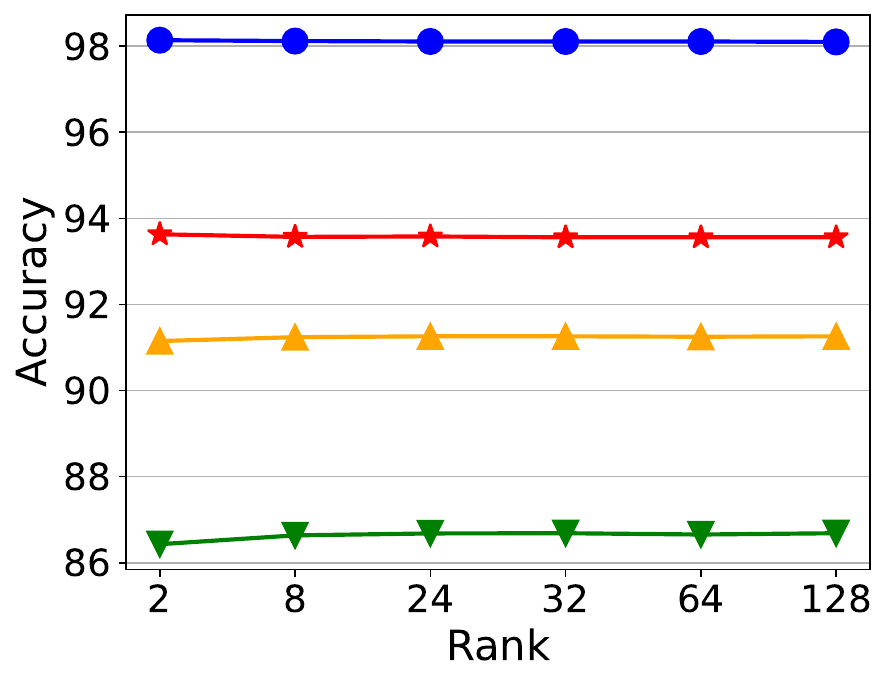}
\includegraphics[width=0.7\columnwidth]{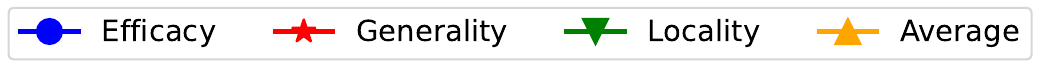}
\caption{Analysis of hyper-parameters $r_o$ and $\alpha_o$. 
}
\label{fig:para}
\end{figure}

In this section, we analyze the impact of different values of alpha $\alpha_o$ and rank $r_o$ in MedLaSA. The results of \textit{Locality} are derived by averaging all its sub-metrics.
As shown in Figure \ref{fig:para}, we observe that as $\alpha_o$ increases, both \textit{Efficacy} and \textit{Generality} also increase. However, \textit{Locality} decreases concurrently. This suggests that the size of $\alpha_o$ significantly influences the model's ability to incorporate new knowledge and its impact on irrelevant knowledge. When selecting $\alpha_o$, the trade-off between the factors must be considered, and the best average result is achieved when $\alpha_o$ equals 24.
On the other hand, as the value of $r_o$ increases, there is no significant change in all metrics. Only when $r_o$ is too small (e.g., equals to 2), does the model's editing \textit{Locality} suffer certain negative effects, which indicates that for single-edit problems, the size of the rank is not a major determinant of the model's performance.

\subsection{Comparison of Editable Weights (RQ5)}

To qualitatively measure the significance of different editable weights and analyze the specific combination of weights that are more suitable for editing in MedLaSA, we conduct a comparison of editing weights in ChatDoctor on the MedFE dataset, as shown in Table \ref{tab:weights}.
For Attn weights, when all four weights are edited together, the learning of editing text is improved, but \textit{Locality} metrics decrease (e.g., from 98.48\% to 90.23\% on CT and from 97.82\% to 86.21\% on TS).
When comparing weights of MLP, $W_{up}$ consistently outperforms $W_{down}$ in all metrics. This suggests that $W_{up}$ may have ability to retain more knowledge and is more suitable for editing medical models.
Moreover, editing MLP weights ($W_{up}$, $W_{down}$, $W_{gate}$) typically leads to significantly higher editing success rate (\textit{Efficacy} > 80\%). 
It is worth noting that when both Attn and MLP weights are simultaneously made trainable for editing, there are additional enhancements in \textit{Efficacy} and \textit{Generality}. However, this comes with significant decreases in \textit{Locality} and \textit{Fluency}. This suggests that incorporating more trainable adapter parameters increases the success rate of medical model editing. Consequently, it leads to a stronger impact on irrelevant information.

\begin{figure}[t]
\centering
\includegraphics[width=1\columnwidth]{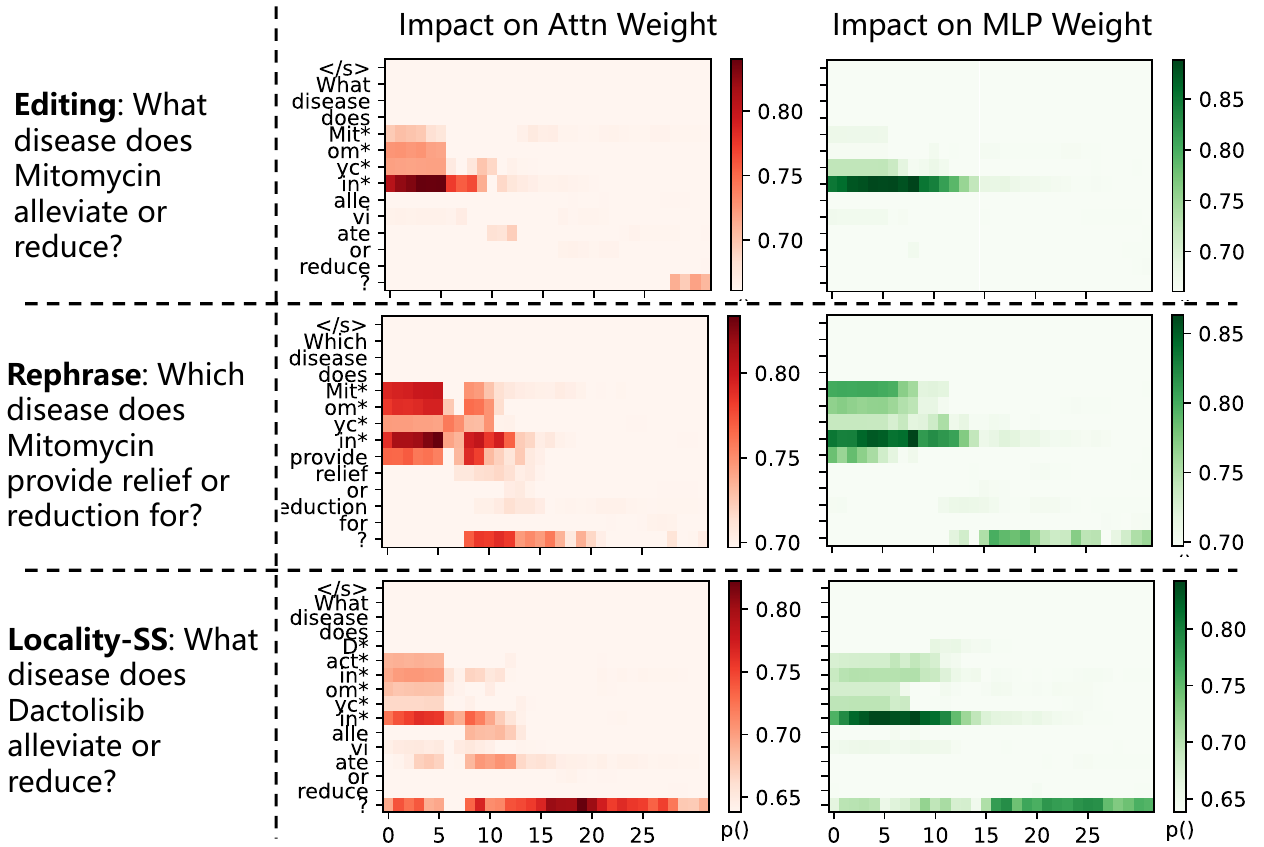}
\caption{Cases of causal tracing on MedCF. The presented red/green heatmaps illustrate the impact of Attn/MLP weight restoration after corrupting input.
}
\label{fig:MedCFCausaltracing}
\end{figure}

\subsection{Case Study (RQ6)}
We present a case study of causal tracing on ChatDoctor network parameters (including Attn and MLP weights) in Figure \ref{fig:MedCFCausaltracing}. These heatmaps provide a visual representation of the association of knowledge with layers and tokens. 
We can observe that editing input and rephrase input (which have the same knowledge but different textual expressions) are strongly associated with layers 5-10. However, unrelated knowledge (\textit{Locality}-SS) focuses on parameters in layers 15-20. This indicates that our MedLaSA is able to achieve a consistent scale set for the same knowledge and a differentiated scale set for unrelated knowledge, ensuring accurate editing of the same knowledge while avoiding impact on unrelated knowledge.


\section{Conclusion}\label{sec:conclusion}

In this paper, we focused on the editing of medical knowledge in LLMs and proposed two preliminary studies: editing factual medical knowledge and editing the explanations of LLMs. Two corresponding benchmarks were constructed to evaluate model editing methods, and more comprehensive and challenging metrics were proposed for \textit{Locality} evaluation.
What is more, we proposed MedLaSA to address the challenges faced in medical model editing due to the specialization and complexity of medical language. 
Extensive experiments conducted on MedCF and MedFE demonstrated the drawbacks of the existing methods and the outperformance of MedLaSA over them. 

\section{ACKNOWLEDGMENTS}
This research was partially supported by CCF-Tencent Open Fund, Tencent Rhino-Bird Focused Research Program, Research Impact Fund (No.R1015-23), CityU - HKIDS Early Career Research Grant (No.9360163), APRC - CityU New Research Initiatives (No.9610565, Start-up Grant for New Faculty of CityU), Hong Kong ITC Innovation and Technology Fund Midstream Research Programme for Universities Project (No.ITS/034/22MS), Hong Kong Environmental and Conservation Fund (No. 88/2022), and SIRG - CityU Strategic Interdisciplinary Research Grant (No.7020046). Additionally, this work was supported in part by the grants from National Natural Science Foundation of China (No.62222213, U23A20319, 62072423).

\bibliographystyle{ACM-Reference-Format}
\bibliography{sample-sigconf}

\appendix









\end{document}